%% file: main.tex
\documentclass{llncs}
\usepackage{amssymb}
\usepackage{amscd}
\usepackage{listings}
\usepackage{wrapfig}
\usepackage{lipsum}
\usepackage{epsfig}
\usepackage{theorem}
\usepackage{paralist}
\usepackage{mathtools}
\usepackage{listings}
\usepackage{algorithm}
\usepackage[noend]{algpseudocode}
\usepackage{subcaption}
\usepackage{graphicx}
\usepackage[usenames,dvipsnames]{color}
\usepackage[hyperfootnotes=false]{hyperref}
\usepackage{theorem,amsmath,amssymb}
\usepackage{graphicx}

\hypersetup{
 colorlinks=false,
 citecolor=Violet,
 linkcolor=Red,
 urlcolor=Blue}

\usepackage[textsize=tiny,textwidth=3cm]{todonotes}

\def\ov#1{\overline{#1}}

\title{An Efficient Summation Algorithm for the Accuracy, Convergence and Reproducibility of Parallel Numerical Methods}

\author{Farah Benmouhoub\inst{1} \and Pierre-Loic Garoche\inst{2} \and Matthieu
  Martel\inst{1,3}}

\institute{
LAMPS Laboratory,
University of Perpignan, France.	\\
\and
ENAC, Toulouse, France.\\
\and
Numalis, Montpellier, France.\\
\email{\inst{1}\{first.last\}@univ-perp.fr, \inst{2}pierre-loic.garoche@enac.fr}
}

\begin{document}

\def\ov#1{\overline{#1}}

\maketitle
\pagestyle{plain}
\begin{abstract}
Nowadays, parallel computing is ubiquitous in several application fields, both in engineering and science. The computations rely on the floating-point arithmetic 
specified by the IEEE754 Standard. In this context, an elementary brick of computation, used everywhere, is the sum of a sequence of numbers.
This sum is subject to many numerical errors in floating-point arithmetic. 
To alleviate this issue, we have introduced a new parallel algorithm for summing a sequence of floating-point numbers. 
This algorithm which scales up easily with the number of processors, adds numbers of the same exponents first.
In this article, our main contribution is an extensive analysis of its efficiency with respect to several properties: accuracy, convergence and reproducibility.
In order to show the usefulness of our algorithm, we have chosen a set of representative numerical methods which are Simpson, Jacobi, LU factorization and the Iterated power method.

\smallskip
\noindent\textbf{Keywords:} floating-point arithmetic, accurate summation, numerical accuracy, numerical methods, convergence, reproducibility.
\end{abstract}

\section{Introduction}\label{intro}
\input{introduction.tex}

\section{Background}\label{background}
This section introduces some useful notions used in the remainder of this article.
Section~\ref{fp} provides some background of the floating-point arithmetic
as defined by the IEEE754 standard~\cite{ieee,goldberg,muller}. Section~\ref{rw} discusses related work.
Our algorithm introduced in~\cite{farah}, is presented in Section~\ref{accalgo}.
\subsection{Floating-Point Arithmetic}\label{fp}
\input{float.tex}
\subsection{Related Work}\label{rw}
\input{related.tex}
\subsection{Accurate Summation Algorithm}\label{accalgo}
\input{accalgo.tex}


\section{Numerical Accuracy}\label{accuracy}
\input{accuracy.tex}

\section{Convergence of Iterative Methods }\label{convergence}
\input{convergence.tex}


\section{Reproducibility }\label{repro}
\input{reproducibility.tex}

\section{Conclusion and Future Work}\label{concl}
\input{conclusion.tex}

\subsubsection*{Acknowledgments}
This work was supported by a regional funding (Region Occitanie) and partially by project ANR-17-CE25-0018 FEANICSES.
\bibliographystyle{plain}
\bibliography{main}  
\end{document}

%% file: introduction.tex
Scientific computing relies heavily on floating-point arithmetic as defined by the IEEE754 Standard~\cite{ieee,goldberg,muller}. It is therefore sensitive to round-off errors, and this problem tends to increase with parallelism.
In floating-point computations, in addition to rounding errors, the order of the computations affects the accuracy of the results.
For example, let us calculate in IEEE754 single precision (Binary32) the sum of
three values $x$, $y$ and $z$, where $x=10^9$, $y=-10^9$ and
$z=10^{-9}$. We obtain
\begin{equation}
\label{addition1}
    ((x+y)+z) = ((10^9-10^9)+10^{-9}) = 10^{-9}, 
\end{equation}
\begin{equation}
\label{addition2}
    (x+(y+z)) = (10^9+(-10^9+10^{-9})) = 0.
\end{equation}
Equations~(\ref{addition1}) and~(\ref{addition2}) show that for the same
mathematical expression, a sum of three operands, different orderings of the computations yield different results.
In the floating-point arithmetic, we note that, for the same values of
$x$, $y$ and $z$, and for the same arithmetic operation, we obtain two
different results because of parsing the three values differently. In fact, many summation algorithms exist in the literature. 
Some of them are based on compensated summation methods~\cite{kahan2,malcolm,rump1,rump2,rump3,rump4} 
with or without the use of the error-free transformations
to compute the error introduced by each accumulation. Others are based on manipulating the exponent
and/or the mantissa of the floating-point numbers in order to split data before starting computations~\cite{demmel1,demmel}.

In a similar approach, we proposed in~\cite{farah} a new algorithm for accurately summing $n$ floating-point numbers. 
This algorithm performs computations only within working precision, requiring only an access to the exponents of the values. The idea is to compute the summands according to their exponents without increasing the complexity.
More precisely, the complexity of the algorithm is linear in the number of elements, just like the naive summation algorithms. The main contribution of the present article is to show that this algorithm improves simultaneously the parallel execution time, the reproducibility and convergence of computations through the increase of their numerical accuracy as follows:
\begin{enumerate}
    \item 
    \emph{numerical accuracy improvement} is illustrated on computations 
of MPI implementations of Simpson's rule and the LU factorization method.
\item \emph{convergence acceleration} is showcased on 
both Jacobi's method and the iterated power method compared to versions of these methods which use a simple summation algorithm. 
Past results~\cite{lopstr15} show that improving the accuracy of computation also leads 
to accelerate the convergence of iterative sequential algorithms. Our motivation is therefore 
to parallelize these two methods focusing first on accuracy and obtaining, as a side effect, a better convergence.
\item Last, \emph{reproducibility of numerical computation in the context of parallel summation} is supported by the reduction on numerical errors.  
Indeed, the combination of the non-associativity of floating-point operations like addition and 
computations done in parallel may affect reproducibility. 
The intuitive solution used to ensure reproducibility is to determine a deterministic order of computation.
Another method is based on reduction or elimination of round-off errors, i.e.
by improving the numerical accuracy of computations that we will further see in this article. This is illustrated on Simpson's rule and a simple matrix multiplication.
\end{enumerate}


A last contribution is an experimental comparison of the execution time of our algorithm with respect to the similar approach proposed by Demmel and Hida~\cite{demmel1}.
Indeed, Demmel and Hida algorithm~\cite{demmel1} has a time complexity of $O(n\log n)$ because of an additional sorting step, compared to our summation algorithm which involve no explicit sorting and has a complexity in $O(n)$. Our solution has a greater space complexity but this can be addressed with sparse datastructures. Execution time of the two approaches are compared in Sections \ref{sec:simpsonrule} and \ref{sec:LU}.

The rest of the paper is organized as follows. Section~\ref{background} recalls elements of floating-point arithmetic and the related work on some existing summation methods proposed to improve the numerical accuracy of computations. We also present our parallel summation algorithm.  
Section~\ref{accuracy} focuses on the improvement of numerical accuracy, based on two experiments, Simpson's rule and a LU factorization. 
Section~\ref{convergence} focuses on the impact of accuracy on the convergence speed. It is based on experiments on Jacobi's method and the Iterated Power Method. Last, we focus on reproducibility in Section~\ref{repro}, based on two experiments: the Simpson's rule and a matrix multiplication.
We conclude in Section~\ref{concl}.

%% file: float.tex
Following the IEEE754 Standard, a floating-point number $x$ in base $\beta$ is defined by
\begin{equation}
x=s\cdot m\cdot\beta^{exp-f+1}
\label{flottants}
\end{equation}
where
\begin{itemize}
\item $s \in \{-1,1\}$ is the sign,
\item $m=d_{0}d_{1}....d_{f-1}$ is the mantissa with digits $0\leq d_{i} <\beta$, $0\leq i\leq f-1$,
\item $f$ is the precision,
\item $exp$ is the exponent with $exp_{min}\leq exp\leq exp_{max}$.
\end{itemize}
The IEEE754 Standard defines binary formats with some particular values for $f$, $exp_{min}$ and $exp_{max}$ which are summarized in Table~\ref{formats}. 
Moreover, the IEEE754 Standard defines four rounding modes for elementary operations over floating-point numbers. These modes are towards $+\infty$, towards $-\infty$, towards zero and towards nearest, denoted by $\circ_{+\infty}$, $\circ_{-\infty}$, $\circ_{0}$ and $\circ_{\sim}$, respectively.

The behavior of the elementary operations $\diamond\in\{+,-,\times,\div\}$ between floating-point numbers is given by
\begin{equation}\label{error}
v_1\diamond_{\circ}v_2=\circ(v_1\diamond v_2)
\end{equation}
where $\circ$ denotes the rounding mode such as $\circ\in\{\circ_{+\infty}, \circ_{-\infty}, \circ_{0},\circ_{\sim}\}$.
By Equation~(\ref{error}), we illustrate that, in floating-point computations, performing an elementary operation $\diamond_\circ$ with rounding mode $\circ$ returns the same result as the one obtained by an exact operation $\diamond$, then rounding the result using $\circ$.
The IEEE754 Standard also specifies how the square root function must be rounded 
in a similar way to Equation~(\ref{error}) but does not specify the round-off of other functions like $\sin$, $\log$, etc.
\begin{table}[h!]
\centering
\begin{tabular}{|cccccc|}
\hline
Format               & \hspace{0.3cm} $\#$total bits & \hspace{0.3cm}$f$ bits & \hspace{0.3cm}$exp$ bits & \hspace{0.3cm}$exp_{min}$ & \hspace{0.3cm}$exp_{max}$\\
\hline
Half precision       & \hspace{0.3cm}$16$ bits      &\hspace{0.3cm} $11$      & \hspace{0.3cm}$5$       & \hspace{0.3cm}$-14$    &\hspace{0.3cm} $+15$ \\
Single precision     & \hspace{0.3cm}$32$ bits      &\hspace{0.3cm} $24$      & \hspace{0.3cm}$8$       & \hspace{0.3cm}$-126$   & \hspace{0.3cm}$+127$ \\
Double precision     & \hspace{0.3cm}$64$ bits      &\hspace{0.3cm} $53$      & \hspace{0.3cm}$11$      & \hspace{0.3cm}$-1122$  & \hspace{0.3cm}$+1223$\\
Quadruple precision  & \hspace{0.3cm}$128$ bits     &\hspace{0.3cm}$113$      & \hspace{0.3cm}$15$      & \hspace{0.3cm}$-16382$ &\hspace{0.3cm}$+16383$ \\
\hline
\end{tabular}
\caption{Binary formats of the IEEE754 Standard.\label{formats}} 
\end{table}
In this article, without loss of generality, we consider that $\beta=2$. 
We assume the rounding mode to the nearest. In floating-point computations, absorption and cancellation may affect the numerical accuracy of computations.
An absorption occurs when adding two floating-point numbers with different orders of magnitude. The small value is absorbed by the large one. A cancellation occurs when two nearly equal numbers are subtracted and the most significant digits cancel each other. 

%% file: related.tex
Summation of floating-point numbers is one of the most basic tasks in numerical analysis.
Research work has focused on improving the numerical accuracy~\cite{demmel1,demmel,kahan2,malcolm,rump1,rump2} or reproducibility~\cite{demmel2} of the computations involving summations.
There are many sequential and parallel algorithms for this task. Surveys of them being presented in~\cite{highama,highamb}.
Floating-point summation is often improved by compensated summation methods~\cite{kahan2,malcolm,rump1,rump2,rump3,rump4} with or without 
the use of error-free transformations to compute the error introduced by each accumulation. We detail some of the compensated summation algorithms further in this section.
The accuracy of summation algorithms can also be improved by manipulating the exponent and the mantissa of the floating-point 
numbers in order to split data before starting computations~\cite{demmel1,demmel}. This approach is the one employed by our algorithm and it is explained in details in Section~\ref{accalgo}.

\textit{Compensated summation methods:}
The idea is to compute the exact rounding error after each addition during computations~\cite{kahan2}. 
Compensated summation algorithms accumulate these errors and add the result to the result of the summation. The compensation process can be applied recursively yielding cascaded compensated algorithm. Malcolm~\cite{malcolm} describes cascading methods based on the limited exponent range of floating-point numbers. 
He defines an extended precision array $e_{i}$ where each component corresponds to an exponent.
To extract and scale the exponent, Malcolm uses an integer division, without requiring the division to be a power of $2$.  
If the extended precision has $53+k$ bits in the mantissa,
then, obviously, no error occurs for up to $2^{k}$ summands and $\sum_{i=1}^{n} p_{i} = \sum_{i=1}^{n} e_{i}$.
The summands $p_{i}$ are added with the respect to decreasing order into the array element corresponding to their exponent. Note that such an algorithm requires twice as much running time compared to our algorithm.

Rump et al.~\cite{rump1,rump2,rump3} proposed several algorithms for summation and dot product of floating point numbers. 
These algorithms are based on iterative application of compensations. 
An extension of the compensation of two floating-point numbers to vectors of arbitrary length is also given and used to compute a result as if computed with twice the working precision.
Various applications of compensated summation method have been proposed~\cite{langlois3,langlois1}. 
Th\'evenoux et al.~\cite{langlois2} implement an automatic code transformation to derive a compensated programs.

Also, we mention the accurate floating-point summation algorithms introduced by Demmel and Hida~\cite{demmel1,demmel}. 
Given two precision $f$ and $F$ with $F>f$, Demmel and Hida's algorithms use a fixed array accumulators $A_0,....A_N$ of precision $F$ 
for summing $n$ floating-point numbers of precision $f$ such that $S=\sum_{i=1}^{n}s_{i}$. 
These algorithms require accessing the exponent field of each $s_{i}$ to decide to which accumulator $A_{j}$ to add it. More precisely, each $A_{j}$ accumulating the sum of the $s_{i}$ where $e$ leading bits are $j$.
Then, these $A_{j}$ are sorted in decreasing order to be summed. Consequently, complexity of these algorithms is equal to $O(n\log n)$, because of the sorting step. \\

\textit{Parallel approaches:}
In addition to the existing sequential algorithms, many other parallel algorithms have been proposed.
Leuprecht and Oberaigner~\cite{leuprechet} describe parallel algorithms for summing floating-point numbers. 
The authors propose a pipeline version of sequential algorithms~\cite{bohlender,pichat1} dedicated to the computation of 
exact rounding summation of floating-point numbers.
In order to ensure the reproducibility, Demmel and Nguyen~\cite{demmel2} introduce a parallel floating-point summation 
method based on a technique called pre-rounding to obtain deterministic rounding errors.
The main idea is to pre-round the  floating-point input values to a common base, according to some boundary, 
so that their sum can later be computed without any rounding error. The error depends only on both input values and the boundary, 
contrary to the intermediate results which depend on the order of computations.


%% file: accalgo.tex
In this section, we describe our summation algorithm introduced in~\cite{farah} for accurately summing $n$ floating-point numbers. 
Our algorithm enjoys the following set of properties. 
First, it improves the numerical accuracy of computations without increasing the cost of complexity compared to the naive algorithm.
Second, it performs all the computations in the original working precision without using accumulators of higher precision.
Last, using this new algorithm, we increase the numerical accuracy and, as a side effect and shown in the next sections, we also improve the execution time and reproducibility of summation.

For the algorithm detailed hereafter, we assume that we have $P$ 
processors and $n$ summands (with $n\gg P$). We assign $n/P$ summands to each processor.
For computing the sum $S=\sum_{i=1}^{n}s_i$, $\forall0\leq i< P$ processor $i$ computes $\sum_{j=i\times n/P}^{(i+1)\times n/P-1} s_{j}$. 
Then a reduction -- a last sum -- is done to compute the final result.
\begin{algorithm}[h!]
\caption{Accurate summation with local sum at each processor\label{algo1}}
\begin{algorithmic}[1]
    \State Initialization of the array sum\_by\_exp
    \State total\_sum=0.0
        
    \For{i=0 : p\_row}  \Comment{p\_row: rows number of each processor}
        \State exp\_s\_{i}=getExp(s[i])+bias \Comment{$getExp$: function used to compute exponent \\\hspace{5.6cm}in base 2}
        \State sum\_by\_exp[exp\_s\_i]=sum\_by\_exp[exp\_s\_i]+s[i]
    \EndFor
\State  local\_sum=0.0
\Comment{Summing locally in order of increasing exponents}
\For{i=0 to exp\_max-exp\_min+1}  
        \State local\_sum=local\_sum+sum\_by\_exp[i]
    \EndFor
\Comment Total sum by processor 0
  \State MPI\_Reduce(\&local\_sum, \&total\_sum, 1, Mpi\_float, Mpi\_sum, 0,Mpi\_comm\_world)
\end{algorithmic}
\end{algorithm}
First of all, Algorithm~\ref{algo1} allocates
an array called $sum\_by\_exp$ which is created and initialized at $0$ for all its elements before starting the summations. The array $sum\_by\_exp$ has $exp\_max-exp\_min+1$ elements.
Let us assume that the exponents of summands range from $exp\_min$ to $exp\_max$. 
The main idea is to sum all the summands whose exponent is $2^{i}$ in the cell $sum\_by\_exp[i+bias]$ such as $bias=-exp\_min$, 
this avoids most absorptions while avoiding to sort explicitly the array.
Let $exp\_(s_{i})$ denote the exponent of $s_{i}$ in base $2$. For computing the sum $S=\sum_{i=1}^{n} s_{i}$, 
each value $s_{i}$ is added to the appropriate cell $sum\_by\_exp[exp\_s_{i}+bias]$ according to its exponent.
For parallelism, each processor has an $sum\_by\_exp$ array to sum locally. 
In order to obtain its local final result, we add these values in increasing
order. Once the final local sums are computed, a reduction is done and the processor receiving the result of the reduction gets the total sum.
To emphasize the property regarding the cost of complexity mentioned above, we note that the cost of Algorithm~\ref{algo1} is $O(n)$, no sorting being performed and the access to the data to be summed being done only once.

Let us mention that another refined implementation of Algorithm~\ref{algo1} has been proposed in~\cite{farah}. This second implementation shares a common idea with Algorithm~\ref{algo1} which is related to the way that the summands are added and differs from Algorithm~\ref{algo1} in that the final sum is not carried out in the same way.
The advantage of this other implementation is that the summation results are more accurate than those of Algorithm~\ref{algo1}. Concerning its drawback, the cost of complexity will be higher in the constant of the $O(n)$ while being linear.
In the following sections, we evaluate the simpler and most cost-effective version of Algorithm~\ref{algo1} regarding accuracy, convergence speed for iterative schema and reproducibility.
We note that our implementations are done in the C programming language with MPI, compiled with MPICC $3.2$, and made them run on an Intel i$5$ with 7.7 GB memory.
Let us also note that for our experiments, we report numerical values relying on thousands separators, typesetting $1234567.89$ as $1,234,567.89$.

%% file: accuracy.tex
\vspace{-0.2cm}
In this section, we first evaluate the numerical accuracy of our algorithm introduced in Section~\ref{accalgo}.
Secondly, we address the issue of the compromise between numerical accuracy and running time of the studied algorithms.
We take into consideration two examples, namely Simpson's rule and the LU factorization method.
For each example, we have implemented two parallel versions, using MPI~\cite{MPI}. The first one, called original program uses simple sums: to sum $n$ values $x_1,....,x_n$ it computes $(((x_1+x_2)+x_3)+....+x_n)$. The second program uses Algorithm~\ref{algo1} and is called, the accurate program. The experiments are carried out for several numbers of processors.

\subsection{Simpson's Rule}
\label{sec:simpsonrule}
\begin{center}
\begin{figure}[h]
\includegraphics[height=6.5cm]{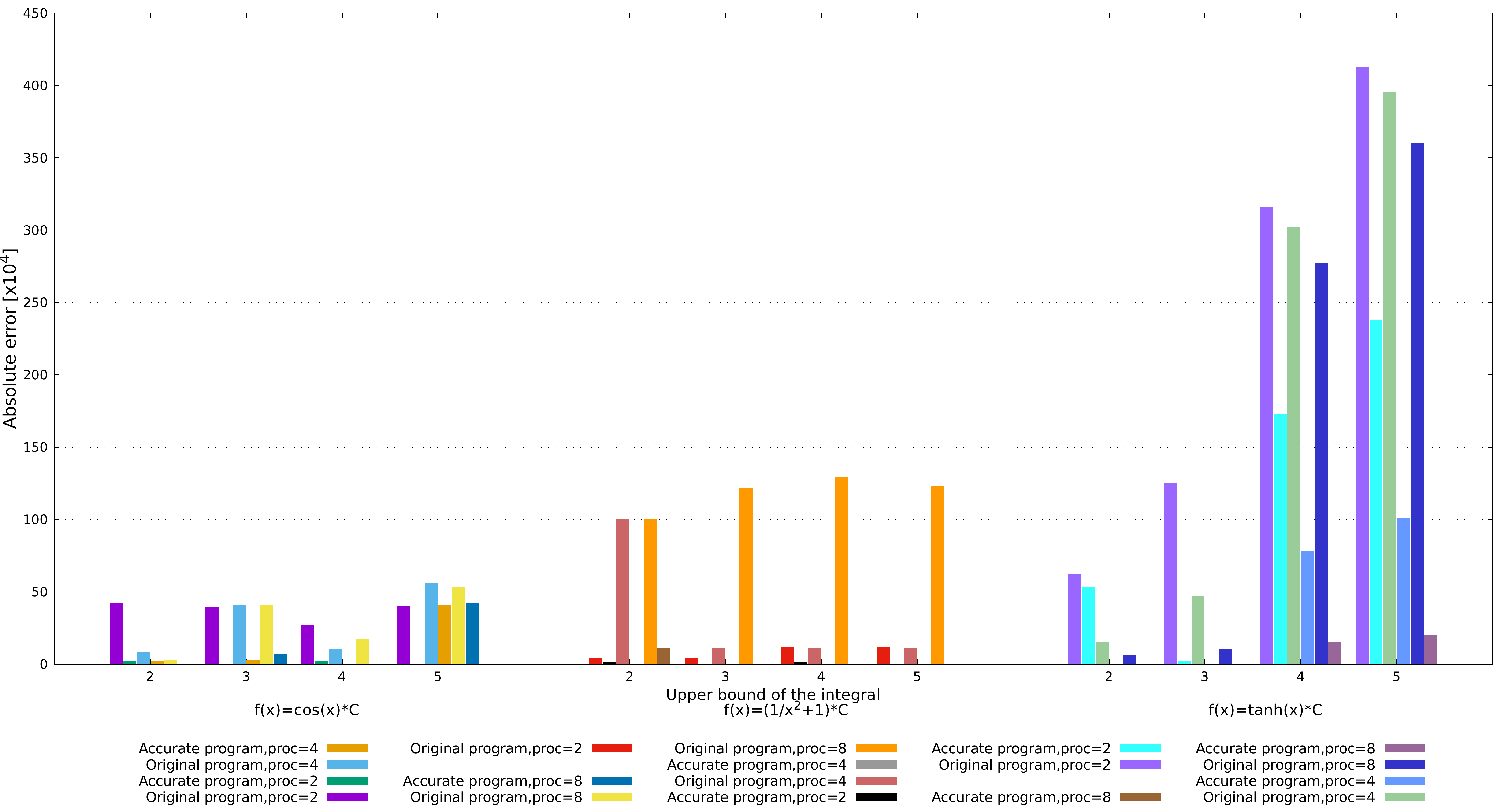}
\caption{The absolute errors between the original program and the accurate one for the integral computation of three different functions $(C\times \cos(x), C\times (1/x^2+1)$ and $C\times \tanh(x))$ with the corresponding analytical result by varying the upper bound of the integral $b=2, 3, 4,5$\label{simpsonacc}.}
\end{figure}
\end{center}

Our first example computes the integral $C\times \int_{0}^{b}f(x)dx$ of mathematical functions $f$ using Simpson's rule.  The Simpson's rule is a numerical method that approximates the value of a definite integral of a function $f$ using quadratic polynomials.
We measure the efficiency of our algorithm on this example by computing the absolute errors between both the results of the original and accurate programs with respect to the analytical solution of the integral, as shown in Figure~\ref{simpsonacc}.
We integrated the following functions $C\times \cos(x)$, $C\times (1/x^{2}+1)$ and $C\times \tanh(x)$ with $C=10^6$, and $b$ ranging in  $[2; 5]$. The number of processors $P$ ranges in $[2; 8]$.
Each processor computes a part of the integral.
\vspace{-0.5cm}
\vspace{-1cm}
For the first experimentation, let us take the function $C\times \cos(x)$ as an example. As it is observed in Figure~\ref{simpsonacc}, the absolute errors of the original program are larger than those of the accurate program of several order of magnitude.
To better illustrate, let us consider the value $3$ of the $x$-axis corresponding to the upper bound of the integral. We notice that the absolute errors of the original program are $392,700.198$, $411,541.22875$ and $414,048.5725$ for $P=2$, $P=4$ and $P=8$, respectively. 
In contrast to the accurate program where the absolute errors computed for the same example are $3,238.3225$, $32,419.6975$ and $77,805.5725$, respectively.
In the same way, we note that the results of the second function $C\times (1/x^{2}+1)$ and the third function $C\times \tanh(x)$ are similar to those of the first function $C\times \cos(x)$.
Moreover, the results of the $C\times \tanh(x)$ function show that the results computed by the original Simpson's rule performed on a large upper bound of the integral are those which have larger absolute errors. Indeed, for $P=2$ the absolute errors computed for the upper bound equal to $2$,$3$,$4$ and $5$ are $621,315.3125$, $1,253,797.375$, $3,166,450.745625$ and $4,130,976.360625$ respectively.\\

\begin{center}
\begin{figure}[h]
\includegraphics[height=6.5cm]{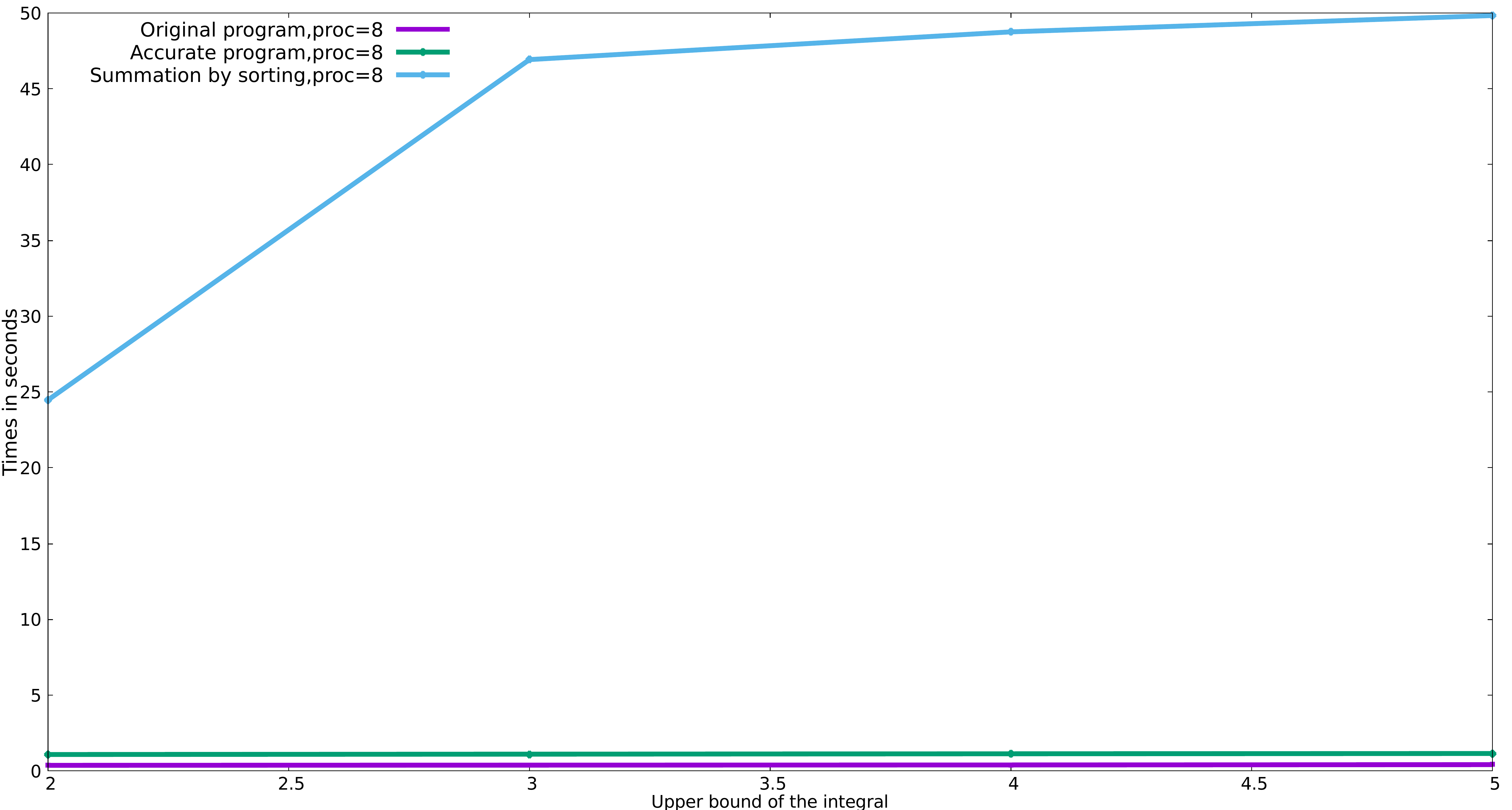}
\caption{ Execution time of the original program, accurate program and summation algorithm by sorting for the integral computation of the function $C\times \tanh(x)$. \label{temps_simp}}
\end{figure}
\end{center}

The second experiment measures the execution time (in seconds) of the original program, accurate program and another program based on sorting. The choice of this last program is motivated by the main idea of Demmel and Hida algoritms~\cite{demmel1}. Let us consider the third function $C\times \tanh(x)$ for $P=8$, Figure~\ref{temps_simp} displays the running time in seconds taken by each program (original program, accurate program and summation by sorting) to compute the integral of this function.
The results show that the summation program based on sorting like Demmel and Hida algorithms~\cite{demmel1} need more time to compute the integral of the function $C\times \tanh(x)$. Contrarily to the summation by sorting program,
our algorithm called accurate program requires much less time and a little more than the original program.
For example, to compute the integral of the function $C\times \tanh(x)$ for $b=2$ it takes $24s$ using the summation by sorting program, while this computation takes only $0.37s$ and $1.08s$ using the original and the accurate programs, respectively.
It is well known that the summations based on sorting performed on the summands before computations are more accurate but these computations take more time ($49s$ for $b=5$) to performed than our accurate algorithm ($1.15s$ for the same value of $b$) where no sorting is performed.

\subsection{LU Factorization}
\label{sec:LU}
Our second example is the parallel LU factorization method. This method consists in rewriting a matrix $A$ as the product of a lower triangular matrix $L$ and an upper triangular matrix $U$ such that $A=LU$. 
The LU factorization method is a very common algorithm which can be used e.g. to solve linear systems or to compute the determinant of a matrix. In the parallel case, the matrix $A$ is divided into blocks of rows and each processor performs its computations on a given block.
For our experiments we generated square matrices of various dimensions $n\in[200,800]$ with increment of $100$. These matrices contain values chosen to introduce ill-conditioned sums~\cite{thevenoux}. In our case, we consider $30\%$ of large values among small and medium. By small, medium and large values we mean respectively, of the order of $10^{-7}$, $10^0$ and $10^{7}$. This is motivated by the IEEE754 single precision arithmetic.
Also, we take vectors $x$ with the same proportions of large values among small and medium as for the matrices.
\begin{center}
\begin{figure}
\includegraphics[height=6.5cm]{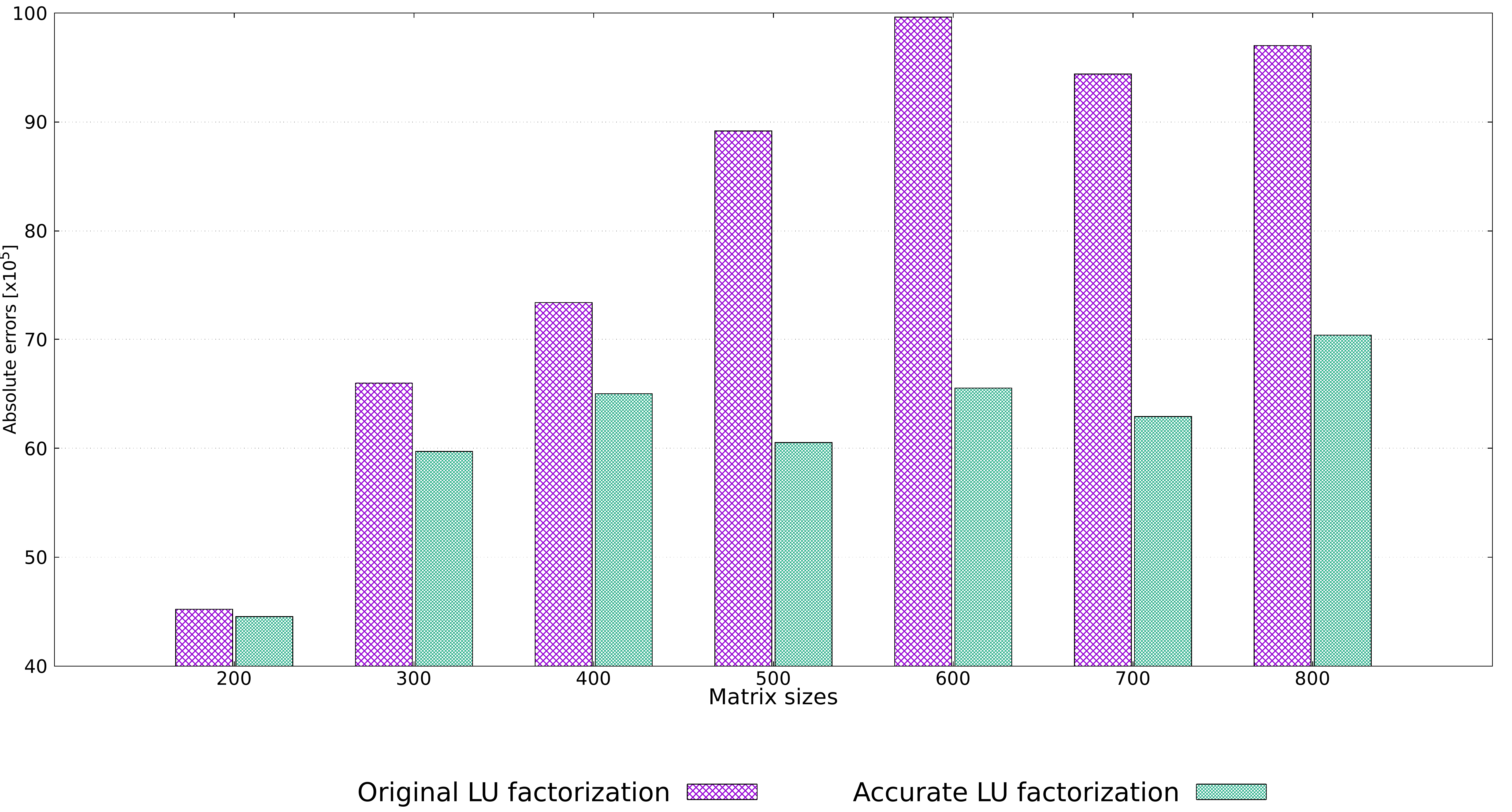}
\caption{The absolute errors of the LU factorization by the original and accurate algorithm for matrix of different sizes\label{luacc}.}
\end{figure}
\end{center}
\vspace{-0.93cm}
The first experimentation consists of comparing the numerical accuracy of the LU factorization carried out using the original and accurate programs. Let us consider a matrix $A$ and vector $x$. We start by computing the solution vector given by $Ax=b$. This vector is considered as the exact solution. Next, we apply the original LU factorization program to the matrix $A$ with $P=16$ processors in order to obtain $L_{orig}$ and $U_{orig}$. In the same way, we factorize the matrix $A$ using the accurate LU factorization into $L_{acc}$ and $U_{acc}$. We compare the new vector solutions $b_{orig}=L_{orig}\times U_{orig}\times x$ and $b_{acc}=L_{acc}\times U_{acc}\times x$ with the exact solution $b$. Figure~\ref{luacc} represents the absolute errors between the computed solutions after factorization and the exact solution.
These experiments show significant improvements: while the difference between the absolute errors of the original and the accurate program are already up to an order of $10^5$ for our smallest matrices ($n=200$), it reaches up to an order of $10^7$ for large ones ($n=600$). 
More precisely, for $n=200$ the absolute errors are $452,1984$ and $445,6448$ for the original and the accurate LU factorization, respectively. We obtain $996,1472$ and $655,3600$ for $n=600$.
Thus, we conclude from this experimentation that the accurate program shows its efficiency in terms of numerical accuracy when we handle large matrices, i.e. when various types of absorptions and cancellation have been introduced.\\

\begin{center}
\begin{figure}[h]
\includegraphics[height=6.5cm]{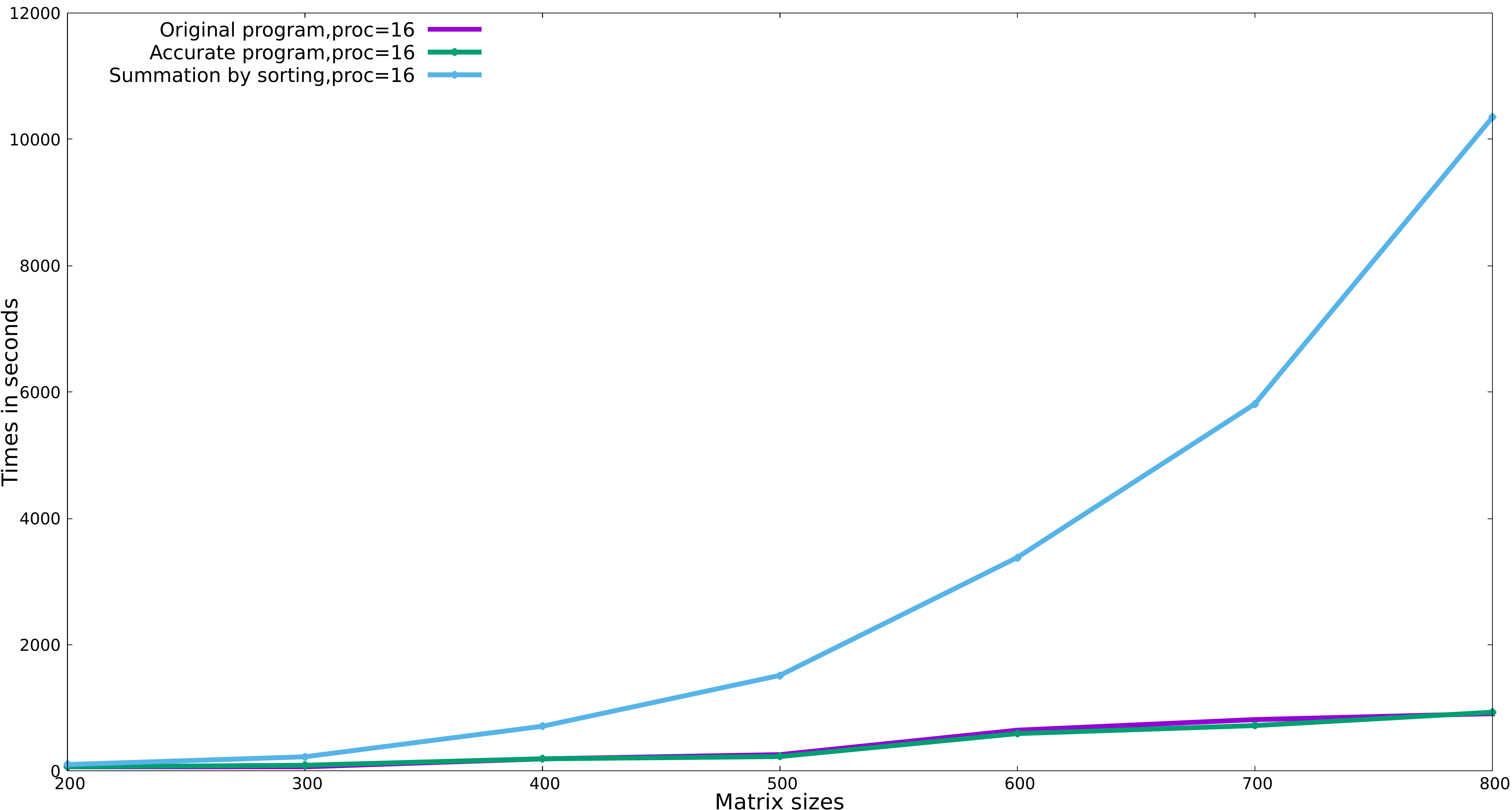}
\caption{Execution time of the original program, accurate program and summation algorithm by sorting of the LU factorization. \label{temps_lu}}
\end{figure}
\end{center}

By the second experimentation, we want to show the running time taken by each LU factorization program for a set of matrices of size varying from $200$ to $800$ with $P=16$. Figure~\ref{temps_lu} summarizes the execution time taken by each algorithm (original program, accurate program and summation by sorting) to compute the LU factorization.
We notice in Figure~\ref{temps_lu} that the summation algorithm based on sorting requires much more time to compute the LU factorization. Besides, accurate program needs much less time than summation algorithm by sorting and a little more than the original algorithm. To better illustrate, let us take a matrix of size $n=300$. We remark that the execution takes only $66s$ and $90s$ with original program and accurate program, respectively, while the summation by sorting program requires $224s$ for the same computation.
Let us also remark that the running time obtained for the summation by sorting program for the large matrices are much larger than those obtained for the smallest one. In fact, the execution time increase from $101s$ to approximately $2h$ for matrices sizes $200$ and $800$, respectively, using the summation algorithm by sorting.

%% file: convergence.tex
In this section, we focus on the impact of accuracy on the number of iterations required by numerical iterative methods to converge. For our experiments, we consider two iterative methods: Jacobi's method and Iterated Power method. 
As in the previous section, we implemented two versions of the same algorithm, the original one and our accurate version.
We observed the impact on the convergence, comparing their respective number of iterations.
\subsection{Jacobi's Method}
\begin{center}
\begin{figure}
\includegraphics[height=6.5cm]{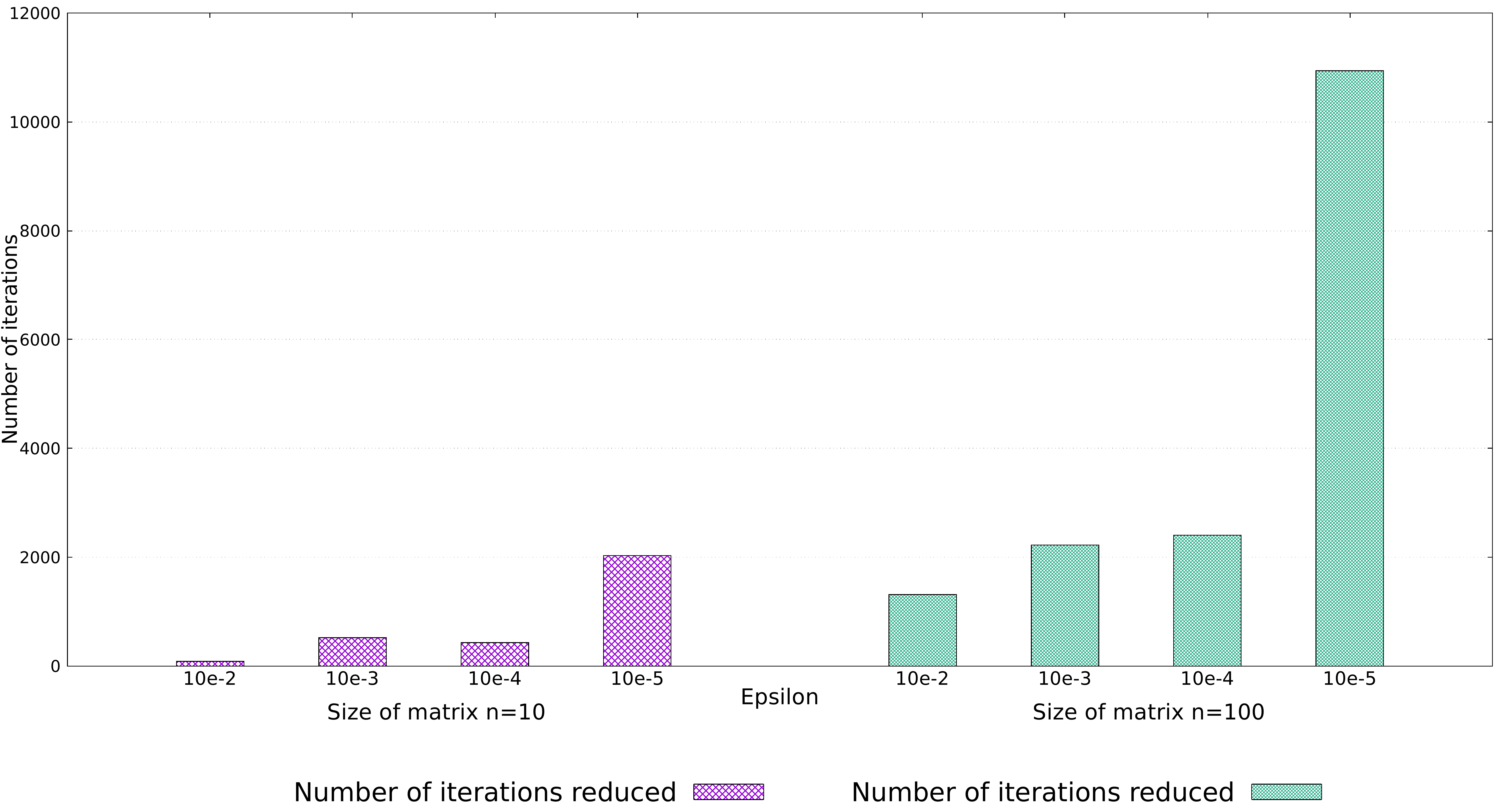}
\caption{Difference between number of iterations necessary for the original and the accurate programs to achieve the convergence of the Jacobi method.\label{jacobi}}
\end{figure}
\end{center}

The Jacobi's method is a well known numerical method used to solve linear systems of the form $Ax=b$.
In this method, an initial guess, an approximate solution $x^{0}$, is selected and is iteratively updated until finding the solution $x^{k}$ of the linear system.
More precisely, this method iterates until $|x_{i}^{(k+1)}-x_i^{k}| < \varepsilon$.
In our case, the parallelization of the Jacobi's method is done according to the row-wise distribution.
Jacobi's method is stable whenever the matrix $A$ is strictly diagonally dominant, i.e. its satisfies the property of Equation~(\ref{stab}). 
\begin{equation}
\label{stab}
\forall i \in 1,\ldots, n,\qquad  |a_{ii}|>\sum_{j \neq i}|a_{ij}|.
\end{equation}
We examine the impact of accuracy on the convergence speed for the systems of sizes $10$ and $100$. While the chosen systems were stable with respect to the sufficient condition of the stability given by Equation~(\ref{stab}), they are close to unstability with $\forall i \in 1,\ldots, n,  |a_{ii}|\approx \sum_{j \neq i}|a_{ij}|.$
Figures~\ref{jacobi} represents the difference between the number of iterations of the original and the accurate programs.
Let us take the first system of size $n=10$. We notice that for various values of $\varepsilon$ varying from $10^{-2}$ to $10^{-5}$, the convergence speed in terms of number of iterations increases from $59$ to $2,029$.
For the second system when $n=100$, we remark that the number of iterations reduced are larger than those computed for the system $n=10$ for the same values of $\varepsilon$. For instance the number of iterations reduced for $\varepsilon=10^{-2}$ and $\varepsilon=10^{-5}$ are $861$ and $10,946$, respectively. 
From these two examples, we conclude that the smallest values of $\varepsilon$ are those which have a large difference between the original and the accurate programs in terms of the number of iterations. Also, for a given value of $\varepsilon$, the accurate program shows its efficiency on the convergence speed on large matrices compared to small ones.
\vspace{-1cm}
\subsection{Iterated Power Method}
\begin{center}
\begin{figure}
\includegraphics[height=6.5cm]{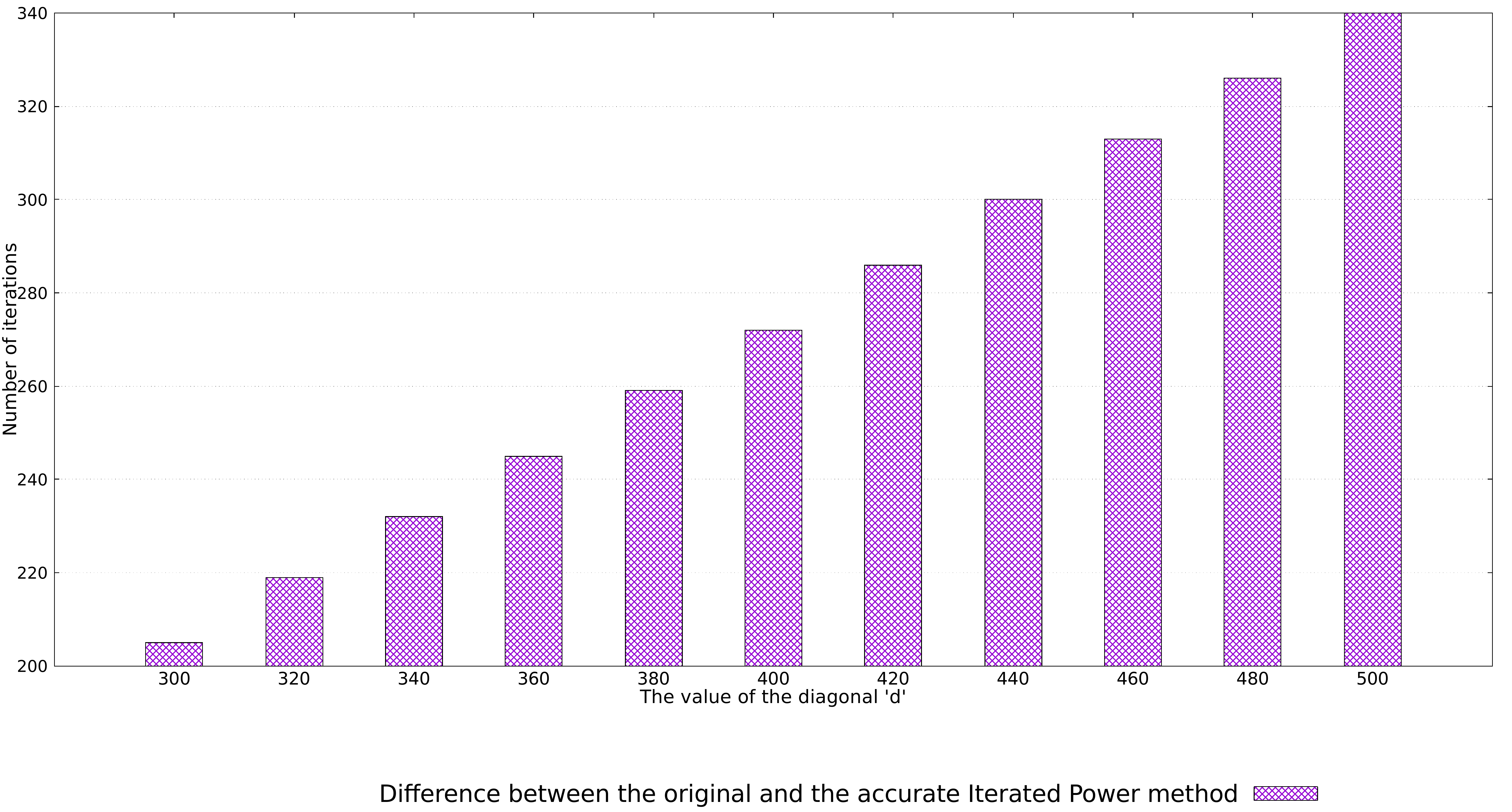}
\caption{Difference between number of iterations of original and accurate Iterated Power method ($n=100$, $d\in[300,500]$ with increment of $20$).\label{power}}
\end{figure}
\end{center}

The Iterative Power method is particularly useful for estimating numerically the largest eigenvalue and its corresponding eigenvector. The idea is to fix an arbitrary initial vector $\textbf{x}^{(0)}$ which contains a single non-zero elements. Then, we build an intermediary vector $\textbf{y}^{(1)}$ such that $\textbf{Ax}^{(0)}=\textbf{y}^{(1)}$. In order to obtain the vector $\textbf{x}^{(1)}$, we renormalize $\textbf{y}^{(1)}$ so that the selected component is again equal to $1$. For the next iteration, we use $\textbf{x}^{(1)}$ as a selected vector.
The iterative process is repeated until convergence. We assume that, the parallelization of the Iterated Power method is done according to the row-wise distribution.
Let us take a square matrix $\textbf{A}$ of the form:
\[
  A=\begin{pmatrix} 
d & a_{12} & \cdots & a_{1j} \\
a_{21} &d &\cdots &a_{2j}  \\
\vdots & & &\vdots \\
a_{i1} & a_{i2}& \cdots & d
\end{pmatrix}
\] 
We assume that $a_{ij}=0.01$ and $d\in[300.0,500.0]$ following the methodology introduced in~\cite{lopstr15}.
Figure~\ref{power} summarizes the difference between the number of iterations of the original and the accurate Iterated Power method.
As it is observed in Figure~\ref{power}, the accurate program accelerates the convergence speed of the Iterative Power method by reducing the number of iterations needed to converge. Indeed, for the matrix size $n=100$ with various values of the diagonal and using $P=4$ we show that the number of iterations reduced increases from $205$ to $340$.

%% file: reproducibility.tex
In this section, we aim at evaluating the efficiency of Algorithm~\ref{algo1} on the improvement of reproducibility.
Figures~\ref{reprosimp} and~\ref{repromat} give results of reproducibility for Simpson's rule and Matrix Multiplication, respectively.
During our experiments, we consider the original and the accurate programs of each method. 
We compare the results of each of them on several processors and their respective results with only one processor.
\subsection{Simpson's Rule}
\begin{center}
\begin{figure}[h]
\includegraphics[height=6.5cm]{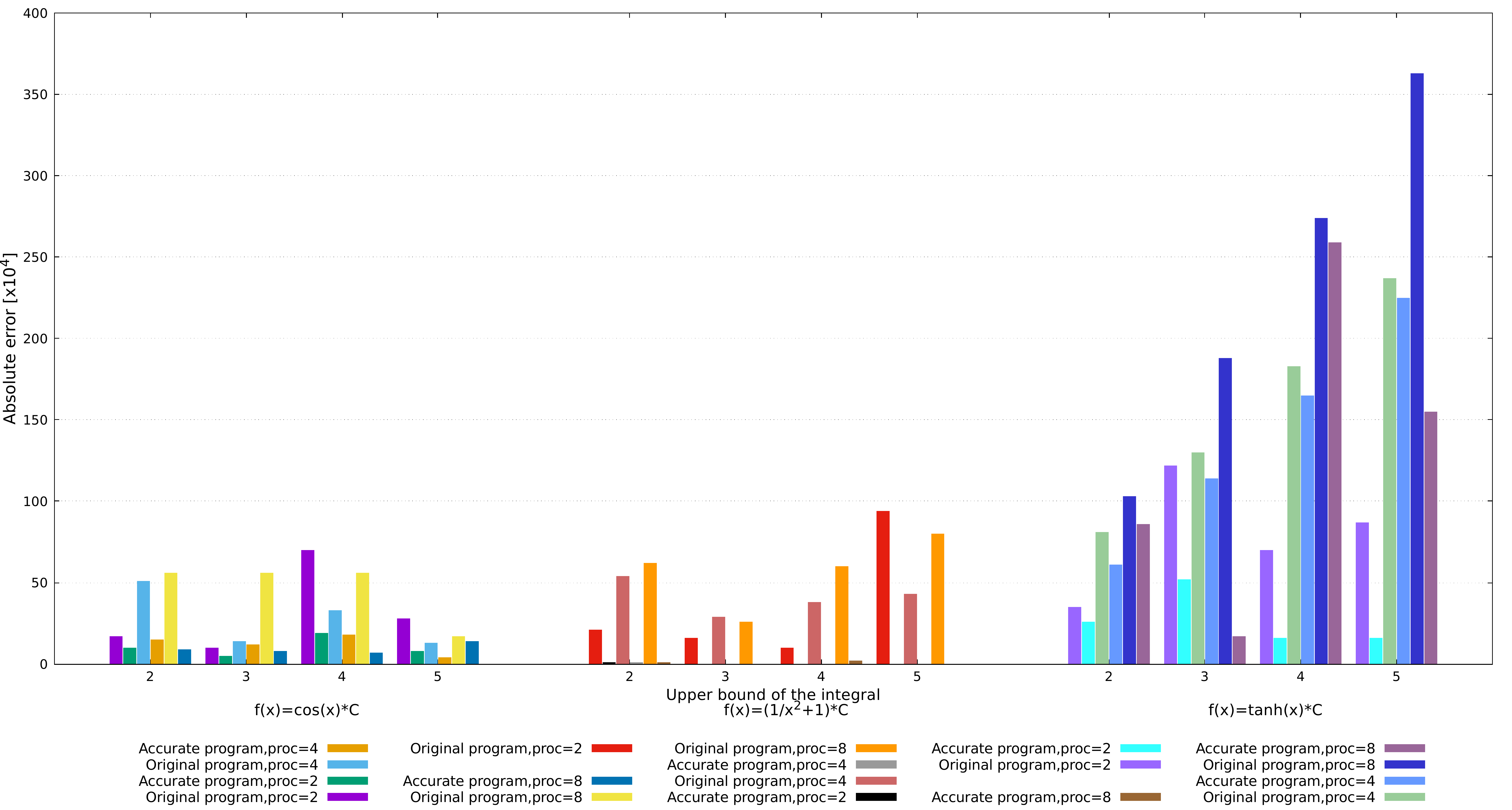}
\caption{The reproducibility of the integral computations using Simpson method of the original program and the accurate one depending on the number of processors.\label{reprosimp}}
\end{figure}
\end{center}

Let's take again the example of Simpson's rule already introduced in Section~\ref{accuracy}. In practice, improving the numerical accuracy often improves reproducibility. 
We measure the efficiency of our algorithm on this example by computing the absolute errors between both the results of the original and the accurate programs by varying the number of processors from $2$ to $8$ and their respective program with only one processor, as shown in Figure~\ref{reprosimp}.
Let us consider several mathematical functions $C\times \cos(x)$, $C\times (1/x^{2}+1)$ and $C\times \tanh(x)$ with $C=10^6$, and $b$ ranging in $[2;5]$.
For example, for $f(x)=C\times \cos(x)$ with $P=2$, the results show that the absolute errors of the original program are larger (between $105,553.117187$ and $703,687.4375$) than those of the accurate program (between $47,938.1875$ and $195,067.296875$) as it is observed in Figure~\ref{reprosimp}.
Also, we can observe in Figure~\ref{reprosimp} that the results of integrals computed by the original program performed on a large number of processors $P=8$ are those which have a larger absolute errors. As an example, the absolute errors computed for the function $C\times \tanh(x)$ with $P=8$ are between $233,122.109375$ and $3,637,171.1875$ while the absolute errors computed for the same example with $P=2$ are between $87,960.921875$ and $1,221,541.8750$.
\vspace{-1cm}
\subsection{Matrix Multiplication}
\begin{center}
\begin{figure}[h]
\includegraphics[height=6.5cm]{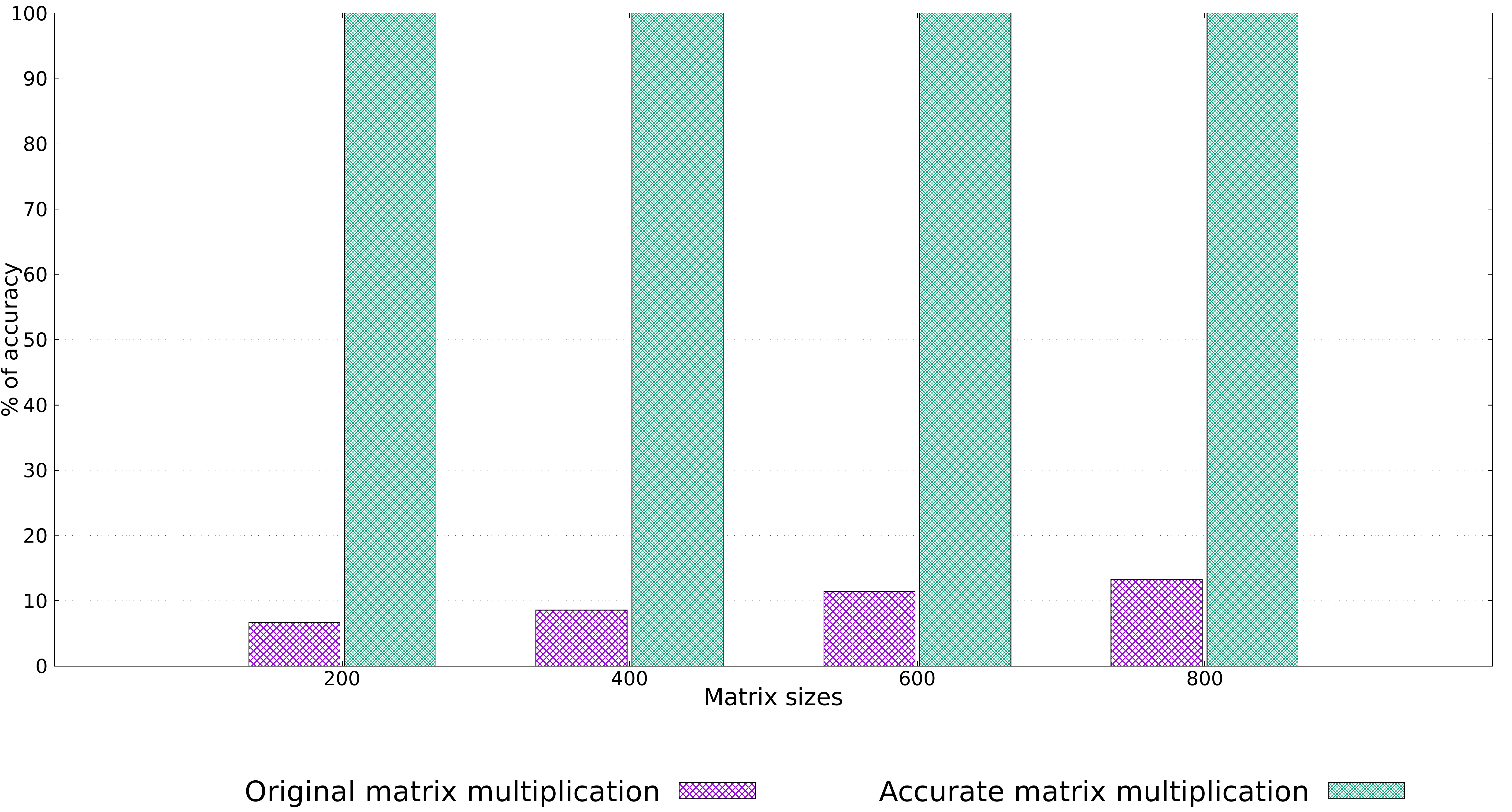}
\caption{The reproducibility of the matrix-matrix multiplication for different size of matrices.\label{repromat}}
\end{figure}
\end{center}

The computation of the matrix-matrix multiplication based on floating-point addition and multiplication which are non-associative operations is prone to accuracy problems.
Moreover, the out of order execution of arithmetic operations on different or even similar parallel architectures are different, reproducibility of the results is not guaranteed.
In this context, we address the problem of reproducibility in the case of matrix multiplication.
To parallelize this method, we assume that each matrix is divided into sub-matrices of the size $n/P$.
For our experiments, we consider square matrices of various dimensions $n\in[200,800]$ with increment of 200.
These matrices contain a variety of floating-point values chosen with difference in magnitude. More precisely, they are made of $50\%$ of large values (of the order of $10^7$) among small (of the order of $10^0$) and medium (of the order of $10^{-7}$). 
Figure~\ref{repromat} represents the percentage of accuracy computed between the original and the accurate programs carried out using $P=8$ and their respective result using only one processor.
The results show that for different matrix sizes, the percentage of accuracy of the original program ranges from $3\%$ to $13\%$.
Contrarily to the original program, the percentage linked to the accurate program described in this article is equal to $100\%$ for each matrix which confirms the usefulness of Algorithm~\ref{algo1}.
\vspace{-1cm}

%% file: conclusion.tex
In this article, we have focused on the impact of accuracy on both the reproducibility and convergence speed of numerical algorithms.
The originality of this article is to study the impact that a new accurate summation algorithm has on the accuracy of numerical methods involving sums. In this work, we have experimented our algorithm on several representative numerical methods by comparing the original and the accurate programs of each of them. The experiments show the usefulness of our algorithm on the improvement of reproducibility. Also, the results obtained show the efficiency of our algorithm to reduce the number of iterations required by numerical iterative methods to converge. More precisely, the accurate program converge more quickly than the original one without loss of accuracy.
In a future work, we would like to refine our algorithm by adding a test phase after line $4$ of Algorithm~\ref{algo1}, i.e. before adding the value $s_i$ to the appropriate cell of the array $sum\_by\_exp$. Indeed, if we have a large set of values to be summed with the same exponent, the result produced can have a larger exponent than the initial one. Therefore, a loss of accuracy can be caused during the computations of the local sums. An interesting perspective consists in feeding our algorithms~\cite{farah} with a static analysis. Our idea is to rely on static analysis to detect the exponents range of a given set of values that will  be summed. Once this range is accurately computed, one can use our summation algorithm and obtain more accurate floating-point results.
Further, it will be interesting to explore the impact of our algorithms in the context of neural networks. 
Our goal is to improve the numerical accuracy of computations by using the accurate summation algorithms~\cite{farah} as a replacement for their summation algorithms.

%% file: main.bbl
\begin{thebibliography}{10}

\bibitem{ieee}
ANSI/IEEE.
\newblock {\em IEEE Standard for Binary Floating-Point Arithmetic}.
\newblock SIAM, 2008.

\bibitem{farah}
Farah Benmouhoub, Pierre-Loic Garoche, and Matthieu Martel.
\newblock {Parallel Accurate and Reproducible Summation}.
\newblock Number~20 in Advances in Intelligent Systems and Computing, London,
  UK, 2021. {Springer}.

\bibitem{bohlender}
Gerd Bohlender.
\newblock Floating-point computation of functions with maximum accuracy.
\newblock {\em IEEE Trans. Comput.}, 26(7):621–632, July 1977.

\bibitem{lopstr15}
Nasrine Damouche, Matthieu Martel, and Alexandre Chapoutot.
\newblock Impact of accuracy optimization on the convergence of numerical
  iterative methods.
\newblock In M.~Falaschi, editor, {\em {LOPSTR} 2015}, volume 9527 of {\em
  Lecture Notes in Computer Science}, pages 143--160. Springer, 2015.

\bibitem{demmel1}
James Demmel and Yozo Hida.
\newblock Accurate floating point summation.
\newblock Technical Report UCB/CSD-02-1180, EECS Department, University of
  California, Berkeley, May 2002.

\bibitem{demmel}
James Demmel and Yozo Hida.
\newblock Accurate and efficient floating point summation.
\newblock {\em SIAM J. Sci. Comput.}, 25(4):1214–1248, April 2003.

\bibitem{demmel2}
James Demmel and Hong~Diep Nguyen.
\newblock Parallel reproducible summation.
\newblock {\em IEEE Transactions on Computers}, 64(7):2060--2070, 2015.

\bibitem{goldberg}
David Goldberg.
\newblock What every computer scientist should know about floating-point
  arithmetic.
\newblock {\em ACM Comput. Surv.}, 23(1):5–48, March 1991.

\bibitem{langlois3}
Stef Graillat, Philippe Langlois, and Nicolas Louvet.
\newblock {Algorithms for Accurate, Validated and Fast Polynomial Evaluation}.
\newblock {\em {Japan Journal of Industrial and Applied Mathematics}},
  26(2-3):191--214, 2009.

\bibitem{langlois1}
Stef Graillat and Val{\'e}rie M{\'e}nissier-Morain.
\newblock {Compensated Horner scheme in complex floating point arithmetic}.
\newblock In {\em {Proceedings, 8th Conference on Real Numbers and Computers}},
  pages 133--146, Santiago de Compostela, Spain, July 2008.

\bibitem{highama}
Nicholas Higham.
\newblock The accuracy of floating point summation.
\newblock {\em SIAM Journal on Scientific Computing}, 14(4):783--799, 1993.

\bibitem{highamb}
Nicholas Higham.
\newblock {\em Accuracy and Stability of Numerical Algorithms}.
\newblock Society for Industrial and Applied Mathematics, USA, 1996.

\bibitem{kahan2}
William Kahan.
\newblock A survey of error analysis.
\newblock In {\em IFIP Congress}, 1971.

\bibitem{thevenoux}
Philippe Langlois, Matthieu Martel, and Laurent Th\'{e}venoux.
\newblock Accuracy versus time: A case study with summation algorithms.
\newblock In {\em Proceedings of the 4th International Workshop on Parallel and
  Symbolic Computation}, PASCO ’10, page 121–130, New York, NY, USA, 2010.
  Association for Computing Machinery.

\bibitem{leuprechet}
H.~Leuprecht and W.~Oberaigner.
\newblock Parallel algorithms for the rounding exact summation of floating
  point numbers.
\newblock {\em Computing}, 28(2):89--104, 1982.

\bibitem{malcolm}
Michael Malcolm.
\newblock On accurate floating-point summation.
\newblock {\em Commun. ACM}, 14(11):731–736, November 1971.

\bibitem{muller}
J.M. Muller, N.~Brisebarre, F.~De Dinechin, C.P. Jeannerod, V.~Lef{\`{e}}vre,
  G.~Melquiond, N.~Revol, D.~Stehl{\'{e}}, and S.~Torres.
\newblock {\em Handbook of Floating-Point Arithmetic}.
\newblock Birkh{\"{a}}user, 2010.

\bibitem{rump1}
Takeshi Ogita, Siegfried Rump, and Shin'ichi Oishi.
\newblock Accurate sum and dot product.
\newblock {\em SIAM Journal on Scientific Computing}, 26(6):1955--1988, 2005.

\bibitem{MPI}
Peter Pacheco.
\newblock {\em An Introduction to Parallel Programming}.
\newblock Morgan Kaufmann Publishers Inc., San Francisco, CA, USA, 1st edition,
  2011.

\bibitem{pichat1}
Mich{\`e}le Pichat.
\newblock Correction d’une somme en arithmetique a virgule flottante.
\newblock {\em Numer. Math.}, 19(5):400–406, October 1972.

\bibitem{rump2}
Siegfried Rump.
\newblock Ultimately fast accurate summation.
\newblock {\em SIAM Journal on Scientific Computing}, 31(5):3466--3502, 2009.

\bibitem{rump3}
Siegfried Rump and Takeshi Ogita.
\newblock Fast high precision summation.
\newblock {\em Nonlinear Theory and Its Applications, IEICE}, 1, 01 2010.

\bibitem{rump4}
Siegfried Rump, Takeshi Ogita, and Shin'ichi Oishi.
\newblock Accurate floating-point summation part {I:} faithful rounding.
\newblock {\em {SIAM} J. Scientific Computing}, 31(1):189--224, 2008.

\bibitem{langlois2}
Laurent Th{\'e}venoux, Philippe Langlois, and Matthieu Martel.
\newblock {Automatic source-to-source error compensation of floating-point
  programs: code synthesis to optimize accuracy and time}.
\newblock {\em {Concurrency and Computation: Practice and Experience}},
  29(7):e3953, 2017.

\end{thebibliography}
